\documentclass{bmvc2k}
\usepackage{graphicx}
\usepackage{subcaption}
\usepackage{booktabs}
\usepackage{wrapfig}
\usepackage{bm}
\usepackage{pifont}
\usepackage{multirow}
\usepackage{graphicx,subfigure}
\usepackage{pifont}
\usepackage{algorithm}
\usepackage{algpseudocode}
\usepackage{booktabs}       
\usepackage{color}
\usepackage{amssymb}
\usepackage{colortbl}
\usepackage[outercaption]{sidecap}
\usepackage[utf8]{inputenc}
\usepackage{fourier} 
\usepackage{array}
\usepackage{makecell}
\usepackage{wrapfig}

\usepackage{caption}

\definecolor{Gray}{gray}{0.90}
\definecolor{white}{rgb}{1.0, 1.0, 1.0}

\definecolor{LightCyan}{RGB}{247, 223, 231}
\newcolumntype{a}{>{\columncolor{LightCyan}}c}
\definecolor{Gray}{gray}{0.90}
\definecolor{LightCyan}{RGB}{247, 223, 231}

\title{Noise-Tolerant Few-Shot Unsupervised Adapter for Vision-Language Models}

\addauthor{Eman Ali}{eman.ali@mbzuai.ac.ae}{1, 2}
\addauthor{Muhammad Haris Khan}{muhammad.haris@mbzuai.ac.ae}{1}

\addinstitution{
 Mohamed Bin Zayed University of Artificial Intelligence,\\
 Abu Dhabi, UAE
}
\addinstitution{
 Alexandria University\\
 Alexandria, Egypt
}

\runninghead{Ali et. al.}{Noise-Tolerant Few-Shot Unsupervised Adapter for VLMs}

\begin{document}

\maketitle

\begin{abstract}
\noindent Recent advances in large-scale vision-language models have achieved impressive performance in various zero-shot image classification tasks. While prior studies have demonstrated significant improvements by introducing few-shot labelled target samples, they still require labelling of target samples, which greatly degrades their scalability and generalizability while handling various visual recognition tasks.
We design NtUA, a Noise-tolerant Unsupervised Adapter that allows the learning of effective target models with few unlabelled target samples. 
NtUA works as a key-value cache that formulates visual features and predicted pseudo-labels of the few unlabelled target samples as key-value pairs. It consists of two complementary designs. The first is adaptive cache formation that combats pseudo-label noises by weighting the key-value pairs according to their prediction confidence. The second is knowledge-guided cache refinement, which refines pair values (i.e., pseudo-labels) and cache weights by leveraging knowledge distillation from large-scale vision language models. Extensive experiments show that NtUA achieves superior performance consistently across multiple widely adopted benchmarks.
\end{abstract}

\section{Introduction}
\label{sec:intro}
The recent progress in large-scale pretrained vision-language models~\cite{radford2021learning,jia2021scaling, yang2022vision} has significantly advanced image-text relationship modelling. One representative work is CLIP~\cite{radford2021learning}\sloppy, which learns image-text relations by jointly training a visual encoder and a linguistic encoder over web-scale paired image-text data. Thanks to the linguistic diversity of the web data, CLIP can be exploited in various image recognition tasks regardless of the number and nature of image classes. The common procedure is to employ CLIP's linguistic encoder to generate text embeddings of pre-defined class names and then match the text embeddings with the features of test images, which are extracted using CLIP's visual encoder in a zero-shot manner. 
\noindent Although pre-trained CLIP has demonstrated significant effectiveness in image classification tasks, its performance depends heavily on the distribution discrepancy between the image-text pairs in CLIP pretraining and specific classification images in various target domains~\cite{radford2021learning, an2024perceptionclip}. Several studies leverage few-shot labelled target samples of each class to adapt the pre-trained CLIP to various target classification tasks to mitigate the inter-domain discrepancy~\cite{gao2024clip, zhang2022tip, zhou2021coop, zhou2022conditional, khattak2023maple, khattak2023self}. Though these approaches have revealed promising results over several few-shot classification benchmarks, they could end up labelling possibly numerous samples for each class of target domains, which significantly affects their scalability, especially while handling large-scale datasets such as ImageNet~\cite{deng2009imagenet} with many image classes. 

Unsupervised learning can adapt pre-trained CLIP to target classification tasks with unlabelled target samples, which offers a more viable alternative to removing the labelling efforts and significantly improves learning scalability.
While recent research has shown some progress in unsupervised adaptation of CLIP across diverse downstream tasks~\cite{huang2022unsupervised, mirza2023lafter, li2022masked, tanwisuth2023pouf}, they often need more substantial computational resources~\cite{li2022masked} or reliance on additional descriptive information for fine-tuning text encoders~\cite{mirza2023lafter}. Moreover, a prevailing assumption in many of these approaches is the availability of abundantly unlabelled samples in the target domain. However, this assumption may need to consistently align with many realistic settings, particularly in contexts where data collection for the target task presents significant challenges, such as medical applications~\cite{zhang2023spectral, gu2022fewshot}, physiological applications~\cite{liu2021metaphys}, or Image-to-Image translation~\cite{9010865}.

\noindent
\begin{minipage}{0.6\textwidth}
Fine-tuning of CLIP model for downstream tasks with a few unlabelled samples is non-trivial and presents significant challenges. First, the model faces the risk of overfitting caused by the scarcity of target domain data~\cite{wu2022style}. Second, the inherent bias of the CLIP model in generating pseudo-labels can result in noisy labels~\cite{li2022masked}. This noise tends to increase as the number of samples per shot decreases (Fig.~\ref{fig:num_samples}), leading to a decline in the adaptation performance (Table.~\ref{sota_ViT32} and Table.~\ref{full}). \emph{To this end, we are the first to explore unsupervised learning for adapting vision-language models towards various downstream classification datasets with a few unlabelled target samples.} 
\end{minipage}%
\hfill
\begin{minipage}{0.35\textwidth}
    \centering
    \includegraphics[width=\textwidth]{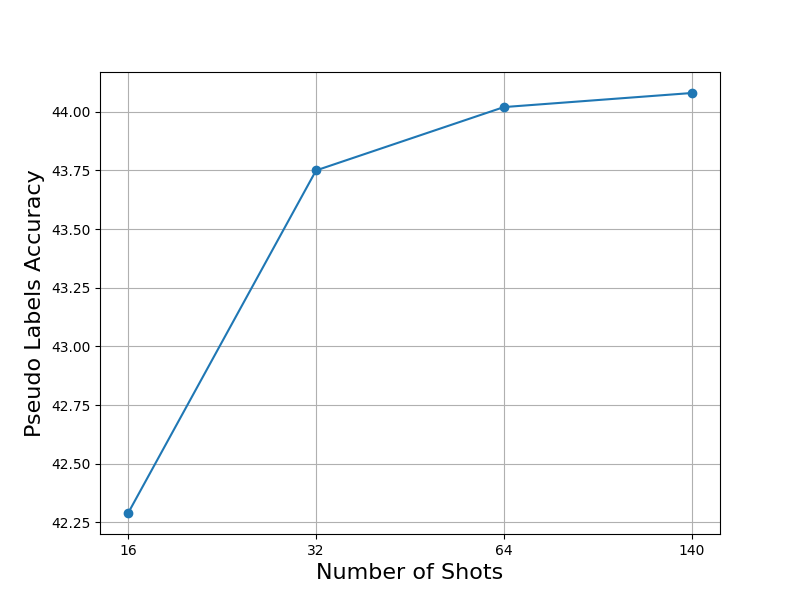}
    \vspace{2pt}
    \captionsetup{font={footnotesize}}
    \captionof{figure}{A comparison of pseudo-label accuracy across different shots within the DTD dataset illustrates that accuracy tends to increase as the number of samples grows.}
    \label{fig:num_samples}
\end{minipage}

\noindent
\begin{minipage}{0.45\textwidth}
We design NtUA, a Noise-tolerant Unsupervised Adapter that enables robust adaptation of pre-trained vision-language models with a few unlabelled target samples. NtUA achieves noise-tolerant adaptation by generating more accurate pseudo-labels for the few unlabelled target samples. NtUA addresses the challenge posed by the higher noise in pseudo-labels when dealing with limited unlabelled data. Inspired by the adapter idea in supervised methods~\cite{grave2017unbounded, khandelwal2019generalization, orhan2018simple,zhang2022tip}, NtUA introduces a weighted key-value cache, which formulates the CLIP-extracted visual features as keys, the predicted pseudo-labels of few target samples as values, and the corresponding pseudo-label confidence as weights of the key-value pairs (Fig.~\ref{fig1}).

\end{minipage}%
\hfill
\begin{minipage}{0.53\textwidth}
    \centering
    \vspace{-10pt}
    \includegraphics[width=\textwidth]{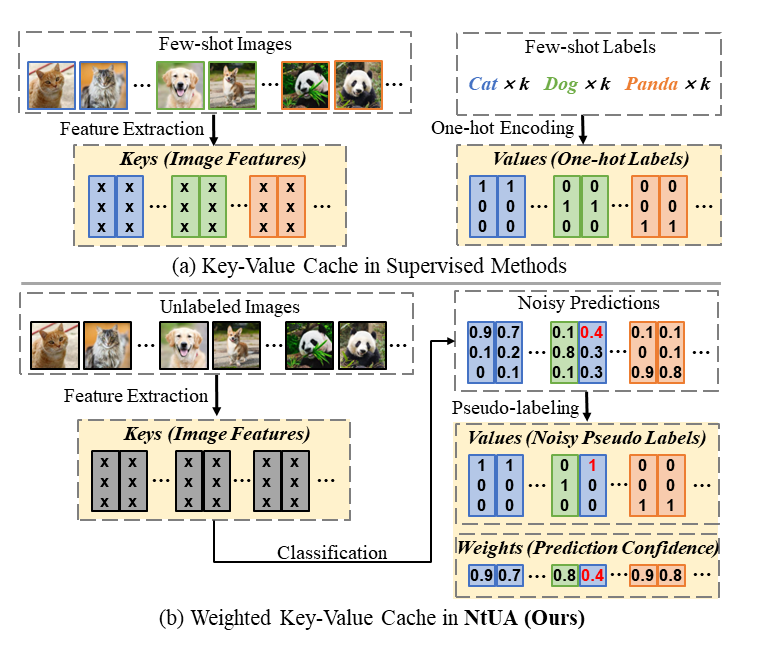}
    \vspace{2pt}
    \captionsetup{font={footnotesize}}
    \captionof{figure}{Unlike key-value cache from labelled samples in supervised methods~\cite{zhang2022tip, orhan2018simple}, we build weighted key-value cache from unlabelled samples, where the cache weights are determined by the confidence of the pseudo-labels predicted by large-scale vision-language models. The adaptive weighting mechanism makes the unsupervised adaptation more tolerant to noisy pseudo-labels.}
    \label{fig1}
\end{minipage}

\noindent Incorporating cache weights significantly enhances NtUA's tolerance to pseudo-label noises, as the prediction confidence is closely correlated with pseudo-label accuracy~\cite{zou2019confidence,wang2020tent}. In addition, we design a knowledge-guided cache refinement technique that effectively improves the quality of the predicted pseudo-labels. This technique leverages CLIP-distilled knowledge, updating pair values and cache weights iteratively. To enhance the robustness of fine-tuning, we incorporate a similarity term between each image and its corresponding class prototype into the loss function, giving higher weight to image-prototype pairs with higher similarity. Extensive experiments show that NtUA is simple but effective, with an average accuracy gain of $6.78\%$ across 11 widely studied datasets.

\section{RELATED WORK}
\noindent\textbf{Cache Model} stores training data features and labels in a key-value format, enabling quick retrieval of relevant information during inference~\cite{vaswani2017attention}. This technique boosts performance across various models, including language models~\cite{grave2017unbounded, merity2016pointer}, vision models~\cite{orhan2018simple}, and vision-language models~\cite{zhang2022tip, zhang2023prompt, zhu2023not}. For example, Tip-Adapter~\cite{zhang2022tip} enhances pre-trained vision-language models with a blended cache model, improving retrieval efficiency. Our novel cache model introduces a weighting mechanism to adapt vision-language models better, prioritizing reliable information for more effective adaptation.

\noindent\textbf{Knowledge Distillation} transfers knowledge from a larger teacher model to a smaller student model, enhancing the student model’s performance~\cite{hinton2015distilling}. Early methods primarily focused on mimicking teacher models’ predicted categorical probabilities~\cite{hinton2015distilling, deepmutuallearning}. Subsequent studies further improved knowledge distillation by transferring teacher knowledge to the student models’ backbone features~\cite{attentiondistillation,detectiondistillation,relational_kd,relational_kd2,peng2019correlation,spkd_gan,zhang2020task}. Unlike most existing studies; we exploit knowledge distillation to learn from powerful vision-language models in the presence of limited unlabelled target samples.

\section{METHODOLOGY}
This section introduces an innovative approach for unsupervised transfer learning, aiming to augment the capability of the pre-trained CLIP model under limited unlabelled target samples. Section~\ref{subsection:A Revisit of Tip-Adapter} presents a concise overview of Tip-Adapter, a closely related study that performs supervised transfer from the pre-trained CLIP model. We then provide details of our proposed Noise-tolerant Unsupervised Adapter (NtUA) in Section~\ref{subsection:Noise-tolerant Unsupervised Adapter}, which is capable of handling a more challenging unsupervised setup that transfers CLIP knowledge with a few unlabelled target samples.

\subsection{A Revisit of Tip-Adapter}
\label{subsection:A Revisit of Tip-Adapter}
Tip-Adapter~\cite{zhang2022tip} is an efficient learning method that adapts the pre-trained CLIP model for supervised few-shot image classification. The task of supervised few-shot image classification involves a labelled dataset $D_{\rm train} = \{(x_i, y_i)\}_{i=1}^{m}$, with an image $x_i$ and its corresponding ground-truth label $y_i$, and the number of samples $m$. $D_{\rm train}$ has $N$-way-$K$-shot, where $N$ represents the number of classes and $K$ represents the number of training examples per class.
For each image $x_{train}$ in $D_{\rm train}$, Tip-Adapter utilizes the visual encoder $E_{v}$ of the pre-trained CLIP model to extract $d$-dimensional L$2$ normalized image features $f_{\rm train} = E_{v}(x_{\rm train})$ and converts its ground-truth label into a $N$-dimensional one-hot vector. 
For all $NK$ training samples, Tip-Adapter extracts $f_{\rm train} \in \mathbb{R}^{1 \times d}$ and $y_{train}\in \mathbb{R}^{1 \times N}$ from each training image to obtain the image features $\mathbf{F}_{\rm train} \in \mathbb{R}^{NK \times d}$ and one-hot vectors $\mathbf{L}_{\rm train} \in \mathbb{R}^{NK \times N}$ for the whole training set.
To build an efficient feature adapter, Tip-Adapter constructs a key-value cache by storing $\mathbf{F}_{\rm train}$ as the keys and $\mathbf{L}_{\rm train}$ as the values.
During inference, the image features $f_\mathrm{test}$ generated from each test image $x_{\rm test} \in D_{\rm test}$ are utilized as a query to retrieve relevant information from the key-value cache. Finally, the prediction logits from Tip-Adapter $P_{\rm TA}$ of a testing image $f_{\rm test}$ is obtained as follows:
\begin{equation}
    P_{\rm TA}(f_{\rm test}) = \alpha \varphi( f_{\rm test} \cdot{} \mathbf{F}^T_{\rm train}) \mathbf{L}_\mathrm{train} + f_{\rm test}  \cdot{} \mathbf{W}^T 
\label{logits_TA}
\end{equation}

\noindent where $\varphi(x) = \exp(-\beta(1 - x))$ is a mapping function defined in~\cite{zhang2022tip}, $\beta$ is a modulating hyperparameter, $\mathbf{W} \in \mathbb{R}^{N\times d}$ denotes the parameters of the textual encoder $E_{t}$ of the pre-trained CLIP, and $\alpha$ refers to a balancing ratio.
The keys in the cache model can be tuned using a loss function defined as follows:
\begin{equation}
        \mathcal{L}_{\text{TA}} (f_{\mathrm{train}}, y_{\mathrm{train}}) = \mathcal{L_{\text{CE}}}(P_{\rm TA}(f_{\mathrm{train}}), \mathbf{L}_{\rm train})
\label{w_logits}
\end{equation}

\subsection{Noise-tolerant Unsupervised Adapter}
\label{subsection:Noise-tolerant Unsupervised Adapter}
\begin{figure}[!t]
 \small \centering
 \begin{minipage}{0.7\textwidth}
  \includegraphics[width=\linewidth]{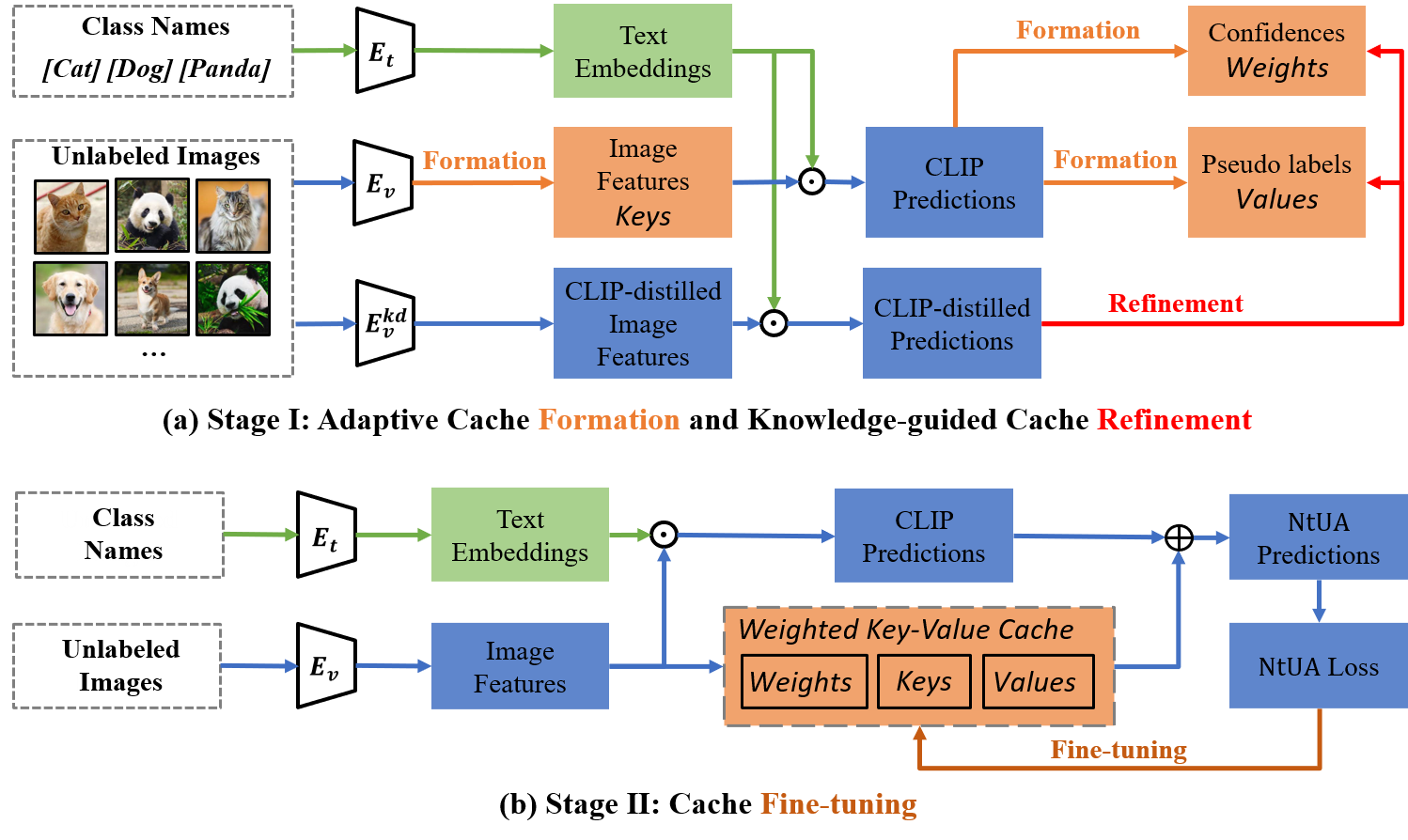}
 \end{minipage}
 \hfill
  \begin{minipage}{1.0\textwidth}
  \captionsetup{font={footnotesize}}
    \caption{The framework of Noise-Tolerant Unsupervised Adapter (NtUA): (a) At Stage I, NtUA first conducts adaptive cache formation by constructing a weighted key-value cache to store the knowledge of few-shot unlabelled target samples and then applies knowledge-guided cache refinement to rectify both cache values and cache weights. In the cache, the image features extracted with CLIP's visual encoder $E_{v}$ serve as the \textit {keys}, the CLIP-predicted pseudo-labels (generated using $E_{v}$ and CLIP's textual encoder $E_{t}$) serve as the \textit {values}. The corresponding pseudo-label prediction confidence serves as the \textit {weights} of the key-value pairs. To perform knowledge-guided cache refinement, NtUA generates CLIP-distilled predictions (with CLIP's visual encoder $E_{v}^{kd}$ and textual encoder $E_{t}$) and leverages such CLIP-distilled knowledge to update both \textit {values} and \textit {weights} in the cache. (b) In Stage II, NtUA updates \textit {the keys} in the constructed \textit {weighted key-value cache} by incorporating knowledge from both the cache and CLIP.}
\label{figure2}
  \end{minipage}
\vspace{-10pt}
\end{figure}

Unlike Tip-Adapter, NtUA aims to transfer CLIP knowledge using only a few unlabelled target samples.
Although Tip-adapter is specifically designed to adapt the CLIP model to target domains with only a few labelled samples, its effectiveness declines when employing limited unlabelled samples for adaptation (see Table.~\ref{ablation_w_KN}).
While CLIP can be utilized to predict pseudo-labels for unlabelled target samples and incorporate these predictions for cache modelling, a significant challenge in this unsupervised scenario arises from inaccuracies in CLIP-generated pseudo-labels. Recent research~\cite{hu2023reclip} has shown limitations in CLIP's visual embedding capabilities, especially for data from less common domains. This weakness often leads to inaccurate pseudo-labels (Table.~\ref{pseudo-labels Quality}).
To address this challenge, NtUA facilitates robust and efficient learning of pseudo-label inaccuracies. As depicted in Fig.~\ref{figure2}, NtUA comprises two crucial stages: \emph{adaptive cache formation} and \emph{knowledge-guided cache refinement}, followed by fine-tuning of the weighted cache. The first stage leverages a larger CLIP model to enhance pseudo-labels' accuracy. The second stage refines the cache model further to improve performance, similar to Tip-Adapter-F~\cite{zhang2022tip}. These stages are detailed in the subsequent subsections, providing insights into their design principles and how they address the challenges of unsupervised CLIP adaptation.

\label{Method_WKC}
\noindent \textbf{Adaptive Cache Formation:} For unsupervised few-shot learning, we possess a set of unlabelled training images denoted as $\mathrm{I}_{NK} = \{x_i\}_{i=1}^{NK}$.
In order to generate pseudo-labels for the unlabelled target data, we employ the visual encoder $E_{v}$ of pre-trained CLIP (ViT-B/32) to extract image features ${f_{\rm train}}$ for all target data. Then, we utilize the CLIP's textual encoder $E_{t}$, which takes a prompt as input, to generate the classifier's weights $\mathbf{W}$ for all the target class names. These weights are applied to the image features, resulting in prediction logits given by $P(f_{train}) = f_{train} \cdot{} \mathbf{W}^T$. 
We then convert these logits into probability distributions using softmax. From these predictions, we derive pseudo-labels as one-hot vectors $\mathbf{\hat{L}_{\rm train}}$ for all unlabelled target images, along with confidence scores $\mathbf{\hat{C}_{\rm train}}$. 

\noindent NtUA incorporates both pseudo-labels $\mathbf{\hat{L}}_{\rm train}$ and their corresponding weights $\mathbf{\hat{C}_{\rm train}}$ in the weighted cache model. It performs inference of new data by combining the predictions of the weighted cache model and the predictions of the CLIP model as follows:
\begin{equation}
\begin{split}
    P_{\text{NtUA}}(f_{\rm train}) 
    & = \alpha \mathbf{\hat{C}_{\rm train}} \varphi(f_{\rm train}  \cdot{} \mathbf{F}^T_{\rm train}) \mathbf{\hat{L}}_{\rm train} + f_{\rm train}  \cdot{} \mathbf{W}^T
\end{split}
\label{NtUA_lgoit}
\end{equation}

\begin{figure}
\begin{tabular}{ccc}
\bmvaHangBox{\fbox{\includegraphics[width=3.5cm]{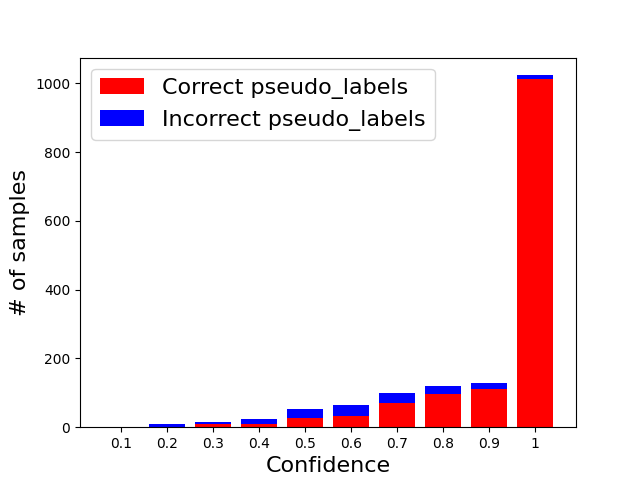}}}&
\bmvaHangBox{\fbox{\includegraphics[width=3.5cm]{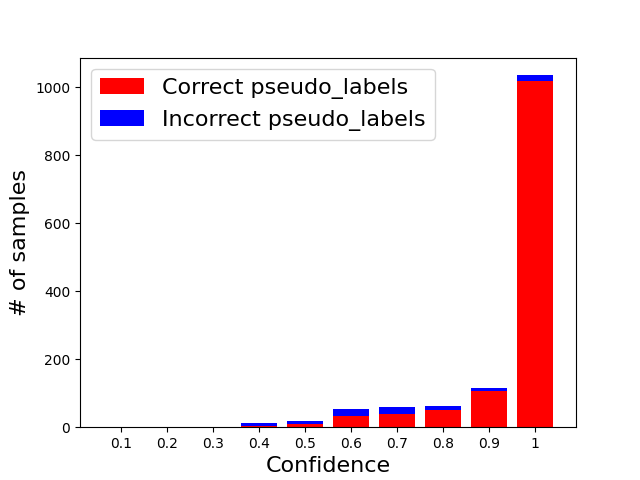}}}&
\bmvaHangBox{\fbox{\includegraphics[width=3.5cm]{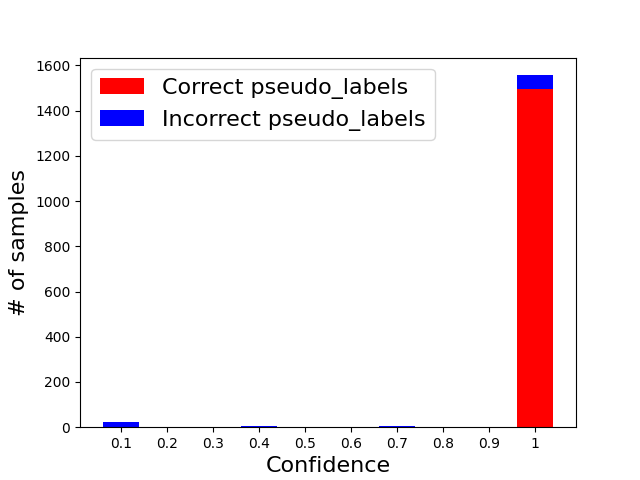}}}\\
\small{(a)} & \small{(b)} & \small{(c)}
\end{tabular}
\vspace{10pt}
\captionsetup{font={footnotesize}}
\caption{The correct and incorrect pseudo-labels distribution based on (a) the confidence generated from ViT-B/32. (b) The confidence generated from ViT-L/14 (c) the prototype-affinity weights $\omega$ generated from ViT-L/14. All experiments are done on a 16-shot sample from the Caltech101 training set.}
\label{correct_vs_incorrect}
\vspace{-12pt}
\end{figure}

\noindent Since the accuracy of pseudo-labels is closely related to the prediction confidence~\cite{zou2019confidence,wang2020tent}, as shown in Fig.~\ref{correct_vs_incorrect}(a), NtUA is more robust to pseudo-label noises by incorporating the prediction confidence $\mathbf{\hat{C}_{\rm train}}$ according to the formulation in Eq.~\ref{NtUA_lgoit}.
To further enhance the quality of both pseudo-labels $\mathbf{\hat{L}}_{\rm train}$ and their corresponding cache weights $\mathbf{\hat{C}_{\rm train}}$. 
We design a novel and effective technique that refines pseudo-labels by leveraging the power of a large CLIP model knowledge. The following subsection provides details about our knowledge-guided cache refinement approach.

\noindent{{\textbf{Knowledge-guided Cache Refinement:}}} \label{Method_PR}To refine the cache model, we leverage the power of another pre-trained CLIP model with a large-scale visual encoder, i.e., ViT-L/14, to effectively enhance the quality of pseudo-labels and their corresponding weights. NtUA initially adopts the visual encoder $\mathrm{E_{v}^{kd}}$ of the large-scale CLIP model to obtain the CLIP-distilled visual features $\mathbf{{F}^{\rm kd}_{\rm train}} = \mathrm{E_{v}^{kd}}(\mathrm{I}_{NK})$, then computes CLIP-distilled predictions $P_{\text{CLIP}}^{kd}$ by multiplying $\mathbf{{F}^{\rm kd}_{\rm train}}$ with the classifier weights $\mathbf{W}$ of CLIP. Then, the CLIP-distilled pseudo-labels $\mathbf{\hat{L}}^{\rm kd}_{\rm train}$ and their corresponding confidence scores $\mathbf{\hat{C}^{\rm kd}_{\rm train}}$ can be obtained by converting the CLIP-distilled predictions $P_{\text{CLIP}}^{kd}$ into one-hot vectors of $N$ dimensions and determining the maximum probabilities among the CLIP-distilled predictions, respectively.

\noindent To perform knowledge-guided cache refinement for the weighted key-value cache, NtUA incorporates the CLIP-distilled pseudo-labels $\mathbf{\hat{L}}^{\rm kd}_{\rm train}$ and their corresponding confidence scores $\mathbf{\hat{C}^{\rm kd}_{\rm train}}$ to update the cache model's pair values and cache weights.
By integrating the updated weighted key-value cache, NtUA enhances the predictions computed in Eq.~\ref{NtUA_lgoit} using the CLIP-distilled pseudo-labels $\mathbf{\hat{L}}^{\rm kd}_{\rm train}$ and the corresponding confidence scores $\mathbf{\hat{C}^{\rm kd}_{\rm train}}$ as follows:
\begin{equation}
\begin{split}
    P_{\text{NtUA}}^{kd}(f_{\rm train}) = 
    & \alpha \mathbf{\hat{C}^{\rm kd}_{\rm train}} \varphi(f_{\rm train} \cdot{} \mathbf{F}^T_{\rm train}) \mathbf{\hat{L}^{\rm kd}_{\rm train}} + f_{\rm train} \cdot{} \mathbf{W}^T
\end{split}
\label{kd_eq}
\end{equation}

\noindent{{\textbf{Weighted Cache Fine-tuning:}}} During fine-tuning, NtUA updates the keys of the cache model by considering the confidence of pseudo-labels generated from the CLIP model architecture. Specifically, prediction logits from the weighted cache are combined with the CLIP's prediction logits to fine-tune the cache model based on the confidence score of the pseudo-labels, as defined in Eq.~\ref{kd_eq}. The loss function in NtUA training can be formulated as follows:
\begin{equation}
        \mathcal{L}_{\text{NtUA}} = \mathcal{L_{\text{CE}}}( P_{\text{NtUA}}^{kd}(f_\mathrm{train}) , \mathbf{\hat{L}}^{\rm kd}_{\rm train}
    )
\label{NtUA_loss}
\end{equation}

\noindent During inference, NtUA leverages the logits obtained from Eq.~\ref{logits_TA} to make accurate predictions. Incorporating the confidence score associated with the pseudo-labels when adjusting the cache model's keys plays a significant role in mitigating the potential adverse effects caused by the presence of inaccurate or noisy pseudo-labels. However, Fig.~\ref{correct_vs_incorrect} (a) and (b) illustrate a key challenge associated with pseudo-labels: a ssomewhattion of the pseudo-labels might be incorrect or assigned with low confidence by the model, even when generated by a large CLIP model like ViT-L/14. While incorporating confidence scores into the adapter helps to some extent, misinterpretations can still occur in specific domains. These misinterpretations can significantly hinder the fine-tuning process.

\noindent Inspired by the few-shot learning paradigm proposed by~\cite{snell2017prototypical, liang2022fewshot, medina2020selfsupervised}, where classification relies on computing the similarity between a data point and class prototypes, we introduce the concept of similarity to prototypes as weights within the loss function. In scenarios involving noisy pseudo-labels with few-shot samples, aggregating image features reduces noise, and the similarity to image prototypes can gauge the reliability of elements belonging to specific classes, as shown in Fig.~\ref {correct_vs_incorrect}(c). We initially aggregate cache values into N class prototypes $\bar{p^{c}}$ through a simple mean of the embeddings for each class: $\bar{p^{c}} = \frac{1}{K}\sum_{i=1}^{K} {f}^{\rm kd}_{\rm train}$. We then derive weights $\omega_{i}^{c}= f_{i}^{c} \cdot \bar{p^{c}}$ by computing the cosine similarity between each cached value and its corresponding image prototype. These weights are referred to as prototype-affinity weights. Then, we employ $\omega$ to adjust the loss function accordingly as follows:
\begin{equation}
        \mathcal{L}_{\text{NtUA}} = \frac{1}{m} \sum_{i=1}^{m} \omega_i \cdot \mathcal{L_{\text{CE}}^{\text{i}}}( P_{\text{NtUA}}^{kd}(f_\mathrm{train}) , \mathbf{\hat{L}}^{\rm kd}_{\rm train})
\label{w_NtUA_loss}
\end{equation}
By prioritizing pseudo-labels with high confidence scores and similarity to image prototypes, we can ensure that the training relies on accurate pseudo-labels, leading to superior overall performance. 

\section{EXPERIMENT}
\noindent \textbf{Datasets:} All the experiments are conducted on 11 widely-used image classification datasets: ImageNet \cite{deng2009imagenet}, Caltech101 \cite{fei2004learning}, DTD \cite{cimpoi2014describing}, EuroSAT \cite{helber2019eurosat}, FGVCAircraft \cite{maji2013fine}, Food101 \cite{bossard2014food}, Flowers102 \cite{nilsback2008automated}, OxfordPets \cite{parkhi2012cats}, SUN397 \cite{xiao2010sun}, StandfordCars \cite{krause20133d}, and UCF101 \cite{soomro2012ucf101}.

\noindent\textbf{Training Setup:} In our experiments, we utilize CLIP~\cite{radford2021learning} as the pre-trained vision-language model with ViT-B/32 visual encoder. 
We employ a large-scale CLIP model with ViT-L/14 as the visual encoder to refine the knowledge-guided cache. Unless otherwise stated, all experiments and the ablation study are conducted using 16 samples per class. 
During fine-tuning, we adopt the Tip-Adapter-F setting proposed by~\cite{zhang2022tip}. Specifically, we freeze the pre-trained CLIP encoders and update only the cache model’s keys. The cache model’s keys undergo fine-tuning for 20 epochs using a batch size of 16, a learning rate set to 0.001, and the AdamW optimizer with a cosine scheduler. Our model training is conducted on a single NVIDIA Quadro RTX6000. Additional information regarding the experiments and the selection of the few unlabelled samples can be found in the supplementary material.

\subsection{Comparison with Baseline Methods}
\label{subsec: Results}
We compare NtUA to four state-of-the-art unsupervised adaptation methods: 1) Zero-shot CLIP~\cite{radford2021learning}; 2) UPL~\cite{huang2022unsupervised}; 3) UPL*~\cite{huang2022unsupervised}; and 4) LaFTer~\cite{mirza2023lafter}. In an unsupervised scenario, we also assess the performance of two CLIP-supervised adaptation models, MaPLe~\cite{khattak2023maple} and PromptSRC~\cite{khattak2023self}.  
We employ CLIP-ViT-L/14 to generate pseudo-labels for MaPLe and PromptSRC. 
The main results of our proposed method obtained from 11 image classification datasets are presented in Table \ref{sota_ViT32}.
Table.~\ref{sota_ViT32} shows that NtUA consistently outperforms the state-of-the-art. NtUA notably outperforms zero-shot CLIP by $+6.78\%$ with minimal training epochs and samples and remains competitive against UPL and UPL$^{*}$ with gains of $+5.89\%$ and $+2.45\%$ respectively. 
This performance improvement over UPL$^{*}$ shows that NtUA's success is not solely reliant on pseudo-label distillation from large models. The confidence and prototype-affinity weights values also play a crucial role.
Moreover, NtUA surpasses LaFTer by $+5.68\%$ solely through integrating these weights during fine-tuning, despite LaFTer's use of a pre-trained GPT-3 classifier. 
Additionally, it outperforms the unsupervised adaptation of two supervised CLIP adaptation methods, MaPLe and PromptSRC, by $+5.58\%$ and $+4.69\%$, respectively. For experiments beyond 16 shots, refer to the supplementary material.

\noindent
\begin{minipage}{0.53\textwidth}
    \vspace{-40pt}
    While achieving high accuracy is crucial, efficiency is equally essential for practical applications. Fig.~\ref{fig:training_time} demonstrates that NtUA outperforms UPL, UPL$^{*}$, LaFTer, MaPLe, and PromptSRC regarding training time. This emphasizes NtUA's capacity to maintain a delicate equilibrium between performance and computational efficiency. 
\end{minipage}%
\hfill
\begin{minipage}{0.45\textwidth}
    \centering
    \includegraphics[width=\textwidth]{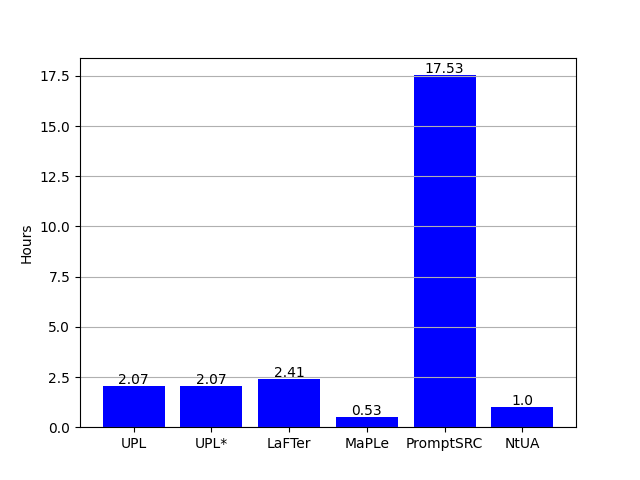}
    \captionsetup{font={footnotesize}}
    \captionof{figure}{Comparison of Training Times: NtUA versus Five Baseline Methods using 16 Unlabelled Samples from ImageNet Dataset}
    \label{fig:training_time}
\end{minipage}

\begin{table}[!t]
  \begin{minipage}{1.0\textwidth}
   \centering \small
  \setlength{\tabcolsep}{3pt}
  \scalebox{0.82}[0.82]{
  \begin{tabular}{c|c|c|c|c|c|c|c|c|c|c|c|c}
  \toprule
  \rowcolor{Gray} 
  \textbf{Methods} & \textbf{ImgNet} & \textbf{Caltech} & \textbf{DTD} & \textbf{ESAT} & \textbf{FGVCA} & \textbf{Food} & \textbf{Flower} & \textbf{OxPets} & \textbf{SUN} & \textbf{StCars} & \textbf{UCF} & \textbf{Average} \\
    \midrule
    CLIP-ViT-B/32   & 63.77 & 91.48 & 44.09 & 45.27 & 19.17 & 80.40 & 66.59& 87.44	& 62.08 & 60.12& 63.47  & 62.17    \\
    \midrule
    UPL  & 58.55 & 90.87 & 48.29 & 58.20 & 18.48 & 79.17 & 69.06 & 84.08 & 65.99 & 55.70 & 65.24  & 63.06   \\
    UPL$^{*}$    & 63.17 & 91.93 & 52.42 & 57.40 & 22.11 & 79.19 & 76.65& 86.37	& \textbf{66.88} & 64.58 & \textbf{70.76}  &  66.50  \\
    LaFTer & 60.41 & 91.60 & 48.93 & 56.06 & 18.30 & 78.72 & 71.49& 85.01	& 62.97 & 57.09 &  65.40 &  63.27  \\
    MaPLe & 62.04 & 92.09 & 46.75 & 52.26 & 20.28 & 80.70 & 71.42 & 85.66 & 63.83 & 56.75 & 65.32  &   63.37 \\						
    PromptSRC & 62.57 & 91.85 & 50.83 & 53.25 & \textbf{22.47} & 79.79 & 73.20 & 85.06 & 63.73	& 57.18 & 66.96 &   64.26    \\
    \rowcolor{LightCyan}
    NtUA (ours)   & \textbf{66.46} & \textbf{94.24} & \textbf{52.96} & \textbf{64.94} & 21.99 & \textbf{81.42} & \textbf{79.90}& \textbf{90.02}	& 66.67 & \textbf{69.25} & 70.61  & \textbf{68.95}   \\
    \midrule
    Supervised   & 68.52 & 95.21 & 69.27 & 85.25 & 36.51 & 82.21 & 93.79& 90.95 & 74.06 & 78.22 &  80.68 & 77.70  \\
  \bottomrule
  \end{tabular}}
  \end{minipage}
  \begin{minipage}{1.0\textwidth}
  \centering 
  \vspace{0.3em}
  \captionsetup{font={footnotesize}}
  \caption{Comparison of NtUA with Five SOTA few-shot adaptation methods over 11 widely adopted image classification benchmarks. All experiments are done with CLIP-ViT-B/32 as a backbone and over 16-shot setups. Supervised denotes supervised fine-tuning using labelled target samples.}
 \label{sota_ViT32}
 \end{minipage}
 \vspace{-1em}
\end{table}

\subsection{Ablation Study}
\label{subsec: Ablation Study}
\noindent\textbf{Different Designs:}\quad
We conduct a series of ablation experiments to examine the specific contributions of each component in NtUA. Four distinct models are tested (see Table.~\ref{ablation_w_KN}):
(1) \textbf{NtUA-(KC)}: Baseline CLIP-ViT-B/32 with Key-value Cache (KC) for unsupervised adaptation of Tip-adapter~\cite{zhang2022tip}.
(2) \textbf{NtUA-(KCR)}: Adds Knowledge-guided Cache Refinement (KCR) with pseudo-labels from CLIP-ViT-L/14.
(3) \textbf{NtUA-(KCR+CKC)}: Integrates Confident weighted Key-value Cache (CKC) on top of \textbf{NtUA-(KCR)}.
(4) \textbf{NtUA-(KCR+CKC+$\omega$)}: Incorporates prototype-affinity weights $\omega$ in the loss function atop \textbf{NtUA-(KCR+CKC)}. 
Compared to \textbf{NtUA-(KC)}, \textbf{NtUA-(KCR)} gains $+3.62\%$ across 11 datasets, showcasing the benefit of using a larger CLIP model for generating pseudo-labels. \textbf{NtUA-(KCR+CKC)} consistently outperforms \textbf{NtUA-(KCR)}, highlighting the value of incorporating confidence scores into cache refinement. Finally, \textbf{NtUA-(KCR+CKC+$\omega$)} surpasses all, achieving $+6.78\%$ improvement over zero-shot CLIP, emphasizing the efficacy of prototype-affinity weights.

\begin{table}[!ht]
  \begin{minipage}{1.0\textwidth}
   \centering \small
  \setlength{\tabcolsep}{3pt}
  \scalebox{0.82}[0.82]{
  \begin{tabular}{c|c|c|c|c|c|c|c|c|c|c|c|c}
  \toprule
  \rowcolor{Gray} 
  Methods & {ImgNet} & {Caltech} & {DTD} & {ESAT} & {FGVCA} & {Food} & {Flower} & {OxPets} & {SUN} & {StCars} & {UCF} & \textbf{Average} \\
    \midrule
    CLIP-ViT-B/32 & 63.77	& 91.48	&44.09	&45.27	&19.17	&80.40	&66.59	&87.44&	62.08	&60.12	&63.47&	62.17\\
    \midrule
    NtUA-(KC)         & 64.88 & 92.74 & 45.21	& 50.05	& 19.23	& 81.01	& 69.14 & 89.40	& 64.78	& 62.03 & 63.89 & 63.85 \\
    NtUA-(KCR)        & 65.58	& 92.98	& 51.24	& 58.91	& \textbf{22.14} & 81.46	& 78.32 & 89.86	& 65.60	& 66.84 & 69.20	& 67.47 \\
    NtUA-(KCR+CKC)    & 65.92	& 92.70	& 52.01 & 60.70	& 22.08	& \textbf{81.52}	& 78.16 & 89.70 & 65.49 & 67.22 & 69.63 & 67.74\\
    \rowcolor{LightCyan}
    NtUA-(KCR+CKC+$\omega$) & \textbf{66.46} & \textbf{94.24} & \textbf{52.96} & \textbf{64.94} & 21.99 & 81.42 & \textbf{79.90}	& \textbf{90.02}	& \textbf{66.67} & \textbf{69.25}	& \textbf{70.61}	& \textbf{68.95} \\
  \bottomrule
  \end{tabular}}
  \end{minipage}
  \begin{minipage}{1.0\textwidth}
  \centering 
  \vspace{0.3em}
  \captionsetup{font={footnotesize}}
  \caption{Ablation studies are conducted across 11 image classification benchmarks using 16-shot unlabelled target samples. KC, KCR, KCR+CKC, and KCR+CKC+$\omega$ correspond to the baseline key-value cache~\cite{zhang2022tip}, our proposed pseudo-label refinement, our proposed weighted key-value cache, and our proposed weighted-based prototypes loss, respectively.}
 \label{ablation_w_KN}
 \end{minipage}
\end{table}

\noindent\textbf{Pseudo-Labels Accuracy:}\quad
We examine the accuracy of pseudo-labels generated using two Vision Transformer (ViT) backbones, ViT-B/32 and ViT-L/14, to assess the effectiveness of knowledge distillation. We focus on understanding how the size of the CLIP model's image encoder affects the quality of pseudo-labels. Employing both backbones to generate pseudo-labels for unlabelled samples across target training datasets, we observe in Table.~\ref{pseudo-labels Quality} that using ViT-L/14 consistently improves pseudo-label quality compared to ViT-B/32. This underscores the importance of model capacity in knowledge distillation tasks, particularly with complex visual data.

\begin{table}[ht]
  \begin{minipage}{1.0\textwidth}
   \centering \small
  \setlength{\tabcolsep}{3pt}
  \scalebox{0.82}[0.82]{
  \begin{tabular}{c|c|c|c|c|c|c|c|c|c|c|c|c}
  \toprule
  \rowcolor{Gray} 
    Methods & {ImgNet} & {Caltech} & {DTD} & {ESAT} & {FGVCA} & {Food} & {Flower} & {OxPets} & {SUN} & {StCars} & {UCF} & \textbf{Average} \\
    \midrule
    ViT-B/32 & 64.78 & 90.84& 44.40 & 45.16 & 18.69 & 79.86 & 64.62& 81.08& 62.38 & 49.78& 59.80 & 60.13 \\
    ViT-L/14  & 77.20 & 95.16& 52.94 & 58.70 & 32.63 & 90.64 & 76.69 & 89.98 & 66.93 & 72.02 & 74.76 &  71.60 \\
  \bottomrule
  \end{tabular}}
  \end{minipage}
  \begin{minipage}{1.0\textwidth}
  \centering 
  \vspace{0.3em}
  \captionsetup{font={footnotesize}}
  \caption{Comparison of Pseudo Label Accuracy for ViT-B/32 and CLIP-ViT-L/14 (Whole Dataset).}
 \label{pseudo-labels Quality}
 \end{minipage}
 \vspace{-1.6em}
\end{table}

\noindent\textbf{NtUA under Transfer Learning:}\quad
We investigate our proposed method, NtUA, in a transfer learning setting, where all unlabelled samples from the target datasets are utilized during training, unlike our default setting, which uses a limited number of images. These labels are then integrated into the construction and fine-tuning of the weighted cache model. Our study utilizes 10 diverse datasets, excluding ImageNet, due to the significant time required to generate pseudo labels and training for its vast size (1.28 million images). By excluding ImageNet, we ensure a more manageable experimental setup while still encompassing a diverse range of datasets for evaluation.
Table.~\ref{full} showcases the detailed results obtained from these experiments. Our proposed NtUA method consistently outperforms state-of-the-art approaches across the majority of the datasets.

\begin{table}[!ht]
\vspace{-.5em}
  \begin{minipage}{1.0\textwidth}
   \centering \small
  \setlength{\tabcolsep}{3pt}
  \scalebox{0.82}[0.82]{
  \begin{tabular}{c|c|c|c|c|c|c|c|c|c|c|c}
  \toprule
  \rowcolor{Gray} 
 Methods & {Caltech} & {DTD} & {ESAT} & {FGVCA} & {Food} & {Flower} & {OxPets} & {SUN} & {StCars} & {UCF} & \textbf{Average} \\
    \midrule
    CLIP-ViT-B/32 & 91.48 & 44.09  & 45.27 & 19.17 & 80.40 & 66.59& 87.44& 62.08 & 60.12& 63.47 & 62.17\\
    \midrule
    UPL  & 91.12 & 45.09 & 51.81 & 17.85 & 80.73 & 68.05 & 83.78 & 63.69 & 50.74 & 60.98 &   61.38 \\
    UPL$^{*}$   & \textbf{92.62} & 52.54 & 62.96 & 22.44 & 82.31 & 77.02 & 86.94 & 64.95 & 61.45 & 69.89 & 67.31  \\
    LaFTer & 92.13 & 50.30 & \textbf{72.60} & 19.11 & 80.21 & 71.66& 84.93& 63.81 & 57.44 & 66.32  & 65.85 \\
    MaPLe  & 91.72 & 48.11 & 55.58 & 12.09 & 81.37 & 70.44 & 85.12 & 64.62 & 55.60 & 66.30 & 63.10\\						
    PromptSRC & 91.97 & 52.07 & 54.79 & \textbf{23.85} & 81.69 & 76.05 & 86.78 & 65.00 & 49.57 & 68.49 & 66.03\\
    \rowcolor{LightCyan}
    NtUA (ours)  & 92.45 & \textbf{57.86} & 67.91 & 22.83 & \textbf{83.71} & \textbf{79.13} & \textbf{91.01}	& \textbf{67.21} &\textbf{71.12} & \textbf{74.52} & \textbf{70.78} \\
    \midrule
    Supervised   & 94.08 & 75.06 & 94.48 & 39.06 & 84.75 & 92.04 & 90.95 & 76.39 & 81.18 &  83.56 &   81.16 \\
  \bottomrule
  \end{tabular}}
  \end{minipage}
  \begin{minipage}{1.0\textwidth}
  \centering 
  \vspace{0.3em}
  \captionsetup{font={footnotesize}}
  \caption{Comparison of NtUA with Unsupervised Adaptation Methods on 10 Image Classification Datasets (Full Training Sets)}
 \label{full}
 \end{minipage}
 \vspace{-1em}
\end{table}

\noindent\textbf{Different Image Encoder Architecture: } We examined NtUA's effectiveness with different CLIP image encoder architectures (ResNet-50). Even with ResNet-50, NtUA maintains its edge over other SOTA, showing only a slight performance decline compared to ViT-B/32. As depicted in Table.~\ref{sota_rn50}, NtUA outperforms zero-shot CLIP with ResNet-50, showing a significant $+7.79\%$ improvement in the 16-shot setting. Furthermore, NtUA consistently surpasses UPL and UPL$^{*}$ in this scenario. A direct comparison involving LaFTer, MaPLe, or PromptSRC was impractical due to their incompatibility with ResNet-50.

\begin{table}[!ht]
\vspace{-.5em}
  \begin{minipage}{1.0\textwidth}
   \centering \small
  \setlength{\tabcolsep}{3pt}
  \scalebox{0.82}[0.82]{
  \begin{tabular}{c|c|c|c|c|c|c|c|c|c|c|c|c}
  \toprule
  \rowcolor{Gray} 
  \textbf{Methods} & \textbf{ImgNet} & \textbf{Caltech} & \textbf{DTD} & \textbf{ESAT} & \textbf{FGVCA} & \textbf{Food} & \textbf{Flower} & \textbf{OxPets} & \textbf{SUN} & \textbf{StCars} & \textbf{UCF} & \textbf{Average} \\
    \midrule
    CLIP-RN50 & 60.32 & 85.76 & 42.79 & 36.15 & 17.01 & 77.38 & 66.10& 85.69& 58.81 & 55.79& 61.96 & 58.89\\
    \midrule
    UPL  & 58.55 &90.43 & 44.44 & 56.58 & 15.48 & 75.64 & 67.97 & 82.09 & 62.36 & 53.58 & 63.76 &  60.99  \\
    UPL$^{*}$   &59.79 & 88.03 & \textbf{51.60} & 58.69& \textbf{21.03} & 75.30 & 76.57 & 84.96 & \textbf{64.32} &  63.04&  68.01 & 64.67 \\
    \rowcolor{LightCyan}
    NtUA (ours)  & \textbf{62.30} & \textbf{92.74} & 50.47 & \textbf{63.85} & 20.64 & \textbf{78.28} & \textbf{78.28} & \textbf{88.01}& 63.64 &\textbf{66.47} & \textbf{68.78}& \textbf{66.68}\\
    \midrule
    Supervised   & 65.26 & 92.94 & 66.90 & 85.07 & 35.94 & 79.33 & 94.36 & 89.42 & 71.38 & 75.45 &  78.75 &  75.89  \\
  \bottomrule
  \end{tabular}}
  \end{minipage}
  \begin{minipage}{1.0\textwidth}
  \centering 
  \vspace{0.3em}
  \captionsetup{font={footnotesize}}
  \caption{Comparison of NtUA with Unsupervised Few-Shot Adaptation (16-Shot, CLIP-RN50).}
 \label{sota_rn50}
 \end{minipage}
 \vspace{-1.8em}
\end{table}

\section{CONCLUSION}
We introduce NtUA, a novel approach called Noise-tolerant Unsupervised Adapter, which enhances target model learning using limited unlabelled samples. NtUA employs a key-value cache representing visual features and predicted pseudo-labels for unlabelled target samples. NtUA uses an adaptive cache formation technique with confidence-based weights to handle noisy pseudo-labels. NtUA also incorporates similarity to image prototypes into the loss function during cache fine-tuning. Additionally, NtUA refines cache values and weights through knowledge distillation from large-scale vision-language models. Experimental evaluations across multiple classification datasets consistently demonstrate NtUA’s superior performance.

\bibliography{egbib}

\begin{thebibliography}{52}
\providecommand{\natexlab}[1]{#1}
\providecommand{\url}[1]{\texttt{#1}}
\expandafter\ifx\csname urlstyle\endcsname\relax
  \providecommand{\doi}[1]{doi: #1}\else
  \providecommand{\doi}{doi: \begingroup \urlstyle{rm}\Url}\fi

\bibitem[An et~al.(2024)An, Zhu, Panaitescu-Liess, Mummadi, and Huang]{an2024perceptionclip}
Bang An, Sicheng Zhu, Michael-Andrei Panaitescu-Liess, Chaithanya~Kumar Mummadi, and Furong Huang.
\newblock Perception{CLIP}: Visual classification by inferring and conditioning on contexts.
\newblock In \emph{The Twelfth International Conference on Learning Representations}, 2024.

\bibitem[Bossard et~al.(2014)Bossard, Guillaumin, and Van~Gool]{bossard2014food}
Lukas Bossard, Matthieu Guillaumin, and Luc Van~Gool.
\newblock Food-101--mining discriminative components with random forests.
\newblock In \emph{European conference on computer vision}, pages 446--461. Springer, 2014.

\bibitem[Cimpoi et~al.(2014)Cimpoi, Maji, Kokkinos, Mohamed, and Vedaldi]{cimpoi2014describing}
Mircea Cimpoi, Subhransu Maji, Iasonas Kokkinos, Sammy Mohamed, and Andrea Vedaldi.
\newblock Describing textures in the wild.
\newblock In \emph{Proceedings of the IEEE Conference on Computer Vision and Pattern Recognition}, pages 3606--3613, 2014.

\bibitem[Deng et~al.(2009)Deng, Dong, Socher, Li, Li, and Fei-Fei]{deng2009imagenet}
Jia Deng, Wei Dong, Richard Socher, Li-Jia Li, Kai Li, and Li~Fei-Fei.
\newblock Imagenet: A large-scale hierarchical image database.
\newblock In \emph{2009 IEEE conference on computer vision and pattern recognition}, pages 248--255. Ieee, 2009.

\bibitem[Fei-Fei et~al.(2004)Fei-Fei, Fergus, and Perona]{fei2004learning}
Li~Fei-Fei, Rob Fergus, and Pietro Perona.
\newblock Learning generative visual models from few training examples: An incremental bayesian approach tested on 101 object categories.
\newblock In \emph{2004 conference on computer vision and pattern recognition workshop}, pages 178--178. IEEE, 2004.

\bibitem[Gao et~al.(2024)Gao, Geng, Zhang, Ma, Fang, Zhang, Li, and Qiao]{gao2024clip}
Peng Gao, Shijie Geng, Renrui Zhang, Teli Ma, Rongyao Fang, Yongfeng Zhang, Hongsheng Li, and Yu~Qiao.
\newblock Clip-adapter: Better vision-language models with feature adapters.
\newblock \emph{International Journal of Computer Vision}, 132\penalty0 (2):\penalty0 581--595, 2024.

\bibitem[Grave et~al.(2017)Grave, Cisse, and Joulin]{grave2017unbounded}
Edouard Grave, Moustapha~M Cisse, and Armand Joulin.
\newblock Unbounded cache model for online language modeling with open vocabulary.
\newblock \emph{Advances in neural information processing systems}, 30, 2017.

\bibitem[Gu et~al.(2022)Gu, Vesal, Kosti, and Maier]{gu2022fewshot}
Mingxuan Gu, Sulaiman Vesal, Ronak Kosti, and Andreas Maier.
\newblock Few-shot unsupervised domain adaptation for multi-modal cardiac image segmentation.
\newblock In \emph{Bildverarbeitung f{\"u}r die Medizin 2022}, pages 20--25, Wiesbaden, 2022. Springer Fachmedien Wiesbaden.
\newblock ISBN 978-3-658-36932-3.

\bibitem[Helber et~al.(2019)Helber, Bischke, Dengel, and Borth]{helber2019eurosat}
Patrick Helber, Benjamin Bischke, Andreas Dengel, and Damian Borth.
\newblock Eurosat: A novel dataset and deep learning benchmark for land use and land cover classification.
\newblock \emph{IEEE Journal of Selected Topics in Applied Earth Observations and Remote Sensing}, 12\penalty0 (7):\penalty0 2217--2226, 2019.

\bibitem[Hinton et~al.(2015)Hinton, Vinyals, and Dean]{hinton2015distilling}
Geoffrey Hinton, Oriol Vinyals, and Jeff Dean.
\newblock Distilling the knowledge in a neural network.
\newblock \emph{arXiv preprint arXiv:1503.02531}, 2015.

\bibitem[Hu et~al.(2024)Hu, Zhang, Xia, Chen, Luo, Sun, Wang, Qiao, Zeng, Sun, et~al.]{hu2023reclip}
Xuefeng Hu, Ke~Zhang, Lu~Xia, Albert Chen, Jiajia Luo, Yuyin Sun, Ken Wang, Nan Qiao, Xiao Zeng, Min Sun, et~al.
\newblock Reclip: Refine contrastive language image pre-training with source free domain adaptation.
\newblock In \emph{Proceedings of the IEEE/CVF Winter Conference on Applications of Computer Vision}, pages 2994--3003, 2024.

\bibitem[Huang et~al.(2022)Huang, Chu, and Wei]{huang2022unsupervised}
Tony Huang, Jack Chu, and Fangyun Wei.
\newblock Unsupervised prompt learning for vision-language models.
\newblock \emph{arXiv preprint arXiv:2204.03649}, 2022.

\bibitem[Jia et~al.(2021)Jia, Yang, Xia, Chen, Parekh, Pham, Le, Sung, Li, and Duerig]{jia2021scaling}
Chao Jia, Yinfei Yang, Ye~Xia, Yi-Ting Chen, Zarana Parekh, Hieu Pham, Quoc~V Le, Yunhsuan Sung, Zhen Li, and Tom Duerig.
\newblock Scaling up visual and vision-language representation learning with noisy text supervision.
\newblock In \emph{ICML}, 2021.

\bibitem[Khandelwal et~al.(2019)Khandelwal, Levy, Jurafsky, Zettlemoyer, and Lewis]{khandelwal2019generalization}
Urvashi Khandelwal, Omer Levy, Dan Jurafsky, Luke Zettlemoyer, and Mike Lewis.
\newblock Generalization through memorization: Nearest neighbor language models.
\newblock \emph{arXiv preprint arXiv:1911.00172}, 2019.

\bibitem[Khattak et~al.(2023{\natexlab{a}})Khattak, Rasheed, Maaz, Khan, and Khan]{khattak2023maple}
Muhammad~Uzair Khattak, Hanoona Rasheed, Muhammad Maaz, Salman Khan, and Fahad~Shahbaz Khan.
\newblock Maple: Multi-modal prompt learning.
\newblock In \emph{Proceedings of the IEEE/CVF Conference on Computer Vision and Pattern Recognition}, pages 19113--19122, 2023{\natexlab{a}}.

\bibitem[Khattak et~al.(2023{\natexlab{b}})Khattak, Wasim, Naseer, Khan, Yang, and Khan]{khattak2023self}
Muhammad~Uzair Khattak, Syed~Talal Wasim, Muzammal Naseer, Salman Khan, Ming-Hsuan Yang, and Fahad~Shahbaz Khan.
\newblock Self-regulating prompts: Foundational model adaptation without forgetting.
\newblock In \emph{Proceedings of the IEEE/CVF International Conference on Computer Vision}, pages 15190--15200, 2023{\natexlab{b}}.

\bibitem[Krause et~al.(2013)Krause, Stark, Deng, and Fei-Fei]{krause20133d}
Jonathan Krause, Michael Stark, Jia Deng, and Li~Fei-Fei.
\newblock 3d object representations for fine-grained categorization.
\newblock In \emph{Proceedings of the IEEE International Conference on computer vision workshops}, pages 554--561, 2013.

\bibitem[Li et~al.(2023)Li, Savarese, and Hoi]{li2022masked}
Junnan Li, Silvio Savarese, and Steven C.~H. Hoi.
\newblock Masked unsupervised self-training for label-free image classification.
\newblock In \emph{ICLR}, 2023.

\bibitem[Li et~al.(2020)Li, Jiang, and Aarabi]{spkd_gan}
Zeqi Li, Ruowei Jiang, and Parham Aarabi.
\newblock Semantic relation preserving knowledge distillation for image-to-image translation.
\newblock In Andrea Vedaldi, Horst Bischof, Thomas Brox, and Jan-Michael Frahm, editors, \emph{Computer Vision -- ECCV 2020}, pages 648--663, Cham, 2020. Springer International Publishing.
\newblock ISBN 978-3-030-58574-7.

\bibitem[Liang et~al.(2022)Liang, Rangrej, Petrovic, and Hassner]{liang2022fewshot}
Kevin~J Liang, Samrudhdhi~B Rangrej, Vladan Petrovic, and Tal Hassner.
\newblock Few-shot learning with noisy labels.
\newblock In \emph{Proceedings of the IEEE/CVF Conference on Computer Vision and Pattern Recognition}, pages 9089--9098, 2022.

\bibitem[Liu et~al.(2019)Liu, Huang, Mallya, Karras, Aila, Lehtinen, and Kautz]{9010865}
Ming-Yu Liu, Xun Huang, Arun Mallya, Tero Karras, Timo Aila, Jaakko Lehtinen, and Jan Kautz.
\newblock Few-shot unsupervised image-to-image translation.
\newblock In \emph{2019 IEEE/CVF International Conference on Computer Vision (ICCV)}, pages 10550--10559, 2019.
\newblock \doi{10.1109/ICCV.2019.01065}.

\bibitem[Liu et~al.(2021)Liu, Jiang, Fromm, Xu, Patel, and McDuff]{liu2021metaphys}
Xin Liu, Ziheng Jiang, Josh Fromm, Xuhai Xu, Shwetak Patel, and Daniel McDuff.
\newblock Metaphys: few-shot adaptation for non-contact physiological measurement.
\newblock In \emph{Proceedings of the conference on health, inference, and learning}, pages 154--163, 2021.

\bibitem[Maji et~al.(2013)Maji, Rahtu, Kannala, Blaschko, and Vedaldi]{maji2013fine}
Subhransu Maji, Esa Rahtu, Juho Kannala, Matthew Blaschko, and Andrea Vedaldi.
\newblock Fine-grained visual classification of aircraft.
\newblock \emph{arXiv preprint arXiv:1306.5151}, 2013.

\bibitem[Medina et~al.(2020)Medina, Devos, and Grossglauser]{medina2020selfsupervised}
Carlos Medina, Arnout Devos, and Matthias Grossglauser.
\newblock Self-supervised prototypical transfer learning for few-shot classification.
\newblock \emph{arXiv preprint arXiv:2006.11325}, 2020.

\bibitem[Merity et~al.(2016)Merity, Xiong, Bradbury, and Socher]{merity2016pointer}
Stephen Merity, Caiming Xiong, James Bradbury, and Richard Socher.
\newblock Pointer sentinel mixture models.
\newblock \emph{arXiv preprint arXiv:1609.07843}, 2016.

\bibitem[Mirza et~al.(2023)Mirza, Karlinsky, Lin, Kozinski, Possegger, Feris, and Bischof]{mirza2023lafter}
M.~Jehanzeb Mirza, Leonid Karlinsky, Wei Lin, Mateusz Kozinski, Horst Possegger, Rogerio Feris, and Horst Bischof.
\newblock Lafter: Label-free tuning of zero-shot classifier using language and unlabeled image collections.
\newblock In \emph{Conference on Neural Information Processing Systems (NeurIPS)}, 2023.

\bibitem[Nilsback and Zisserman(2008)]{nilsback2008automated}
Maria-Elena Nilsback and Andrew Zisserman.
\newblock Automated flower classification over a large number of classes.
\newblock In \emph{2008 Sixth Indian Conference on Computer Vision, Graphics \& Image Processing}, pages 722--729. IEEE, 2008.

\bibitem[Orhan(2018)]{orhan2018simple}
Emin Orhan.
\newblock A simple cache model for image recognition.
\newblock \emph{Advances in Neural Information Processing Systems}, 31, 2018.

\bibitem[Park et~al.(2019)Park, Kim, Lu, and Cho]{relational_kd}
Wonpyo Park, Dongju Kim, Yan Lu, and Minsu Cho.
\newblock Relational knowledge distillation.
\newblock In \emph{Proceedings of the IEEE/CVF conference on computer vision and pattern recognition}, pages 3967--3976, 2019.

\bibitem[Parkhi et~al.(2012)Parkhi, Vedaldi, Zisserman, and Jawahar]{parkhi2012cats}
Omkar~M Parkhi, Andrea Vedaldi, Andrew Zisserman, and CV~Jawahar.
\newblock Cats and dogs.
\newblock In \emph{2012 IEEE conference on computer vision and pattern recognition}, pages 3498--3505. IEEE, 2012.

\bibitem[Peng et~al.(2019)Peng, Jin, Liu, Li, Wu, Liu, Zhou, and Zhang]{peng2019correlation}
Baoyun Peng, Xiao Jin, Jiaheng Liu, Dongsheng Li, Yichao Wu, Yu~Liu, Shunfeng Zhou, and Zhaoning Zhang.
\newblock Correlation congruence for knowledge distillation.
\newblock In \emph{Proceedings of the IEEE/CVF International Conference on Computer Vision}, pages 5007--5016, 2019.

\bibitem[Radford et~al.(2021)Radford, Kim, Hallacy, Ramesh, Goh, Agarwal, Sastry, Askell, Mishkin, Clark, et~al.]{radford2021learning}
Alec Radford, Jong~Wook Kim, Chris Hallacy, Aditya Ramesh, Gabriel Goh, Sandhini Agarwal, Girish Sastry, Amanda Askell, Pamela Mishkin, Jack Clark, et~al.
\newblock Learning transferable visual models from natural language supervision.
\newblock In \emph{International conference on machine learning}, pages 8748--8763. PMLR, 2021.

\bibitem[Snell et~al.(2017)Snell, Swersky, and Zemel]{snell2017prototypical}
Jake Snell, Kevin Swersky, and Richard Zemel.
\newblock Prototypical networks for few-shot learning.
\newblock \emph{Advances in neural information processing systems}, 30, 2017.

\bibitem[Soomro et~al.(2012)Soomro, Zamir, and Shah]{soomro2012ucf101}
Khurram Soomro, Amir~Roshan Zamir, and Mubarak Shah.
\newblock A dataset of 101 human action classes from videos in the wild.
\newblock \emph{Center for Research in Computer Vision}, 2\penalty0 (11):\penalty0 1--7, 2012.

\bibitem[Tanwisuth et~al.(2023)Tanwisuth, Zhang, Zheng, He, and Zhou]{tanwisuth2023pouf}
Korawat Tanwisuth, Shujian Zhang, Huangjie Zheng, Pengcheng He, and Mingyuan Zhou.
\newblock Pouf: Prompt-oriented unsupervised fine-tuning for large pre-trained models.
\newblock In \emph{International Conference on Machine Learning}, pages 33816--33832. PMLR, 2023.

\bibitem[Tung and Mori(2019)]{relational_kd2}
Frederick Tung and Greg Mori.
\newblock Similarity-preserving knowledge distillation.
\newblock In \emph{Proceedings of the IEEE/CVF international conference on computer vision}, pages 1365--1374, 2019.

\bibitem[Vaswani et~al.(2017)Vaswani, Shazeer, Parmar, Uszkoreit, Jones, Gomez, Kaiser, and Polosukhin]{vaswani2017attention}
Ashish Vaswani, Noam Shazeer, Niki Parmar, Jakob Uszkoreit, Llion Jones, Aidan~N Gomez, {\L}ukasz Kaiser, and Illia Polosukhin.
\newblock Attention is all you need.
\newblock In \emph{Advances in neural information processing systems}, pages 5998--6008, 2017.

\bibitem[Wang et~al.(2020)Wang, Shelhamer, Liu, Olshausen, and Darrell]{wang2020tent}
Dequan Wang, Evan Shelhamer, Shaoteng Liu, Bruno Olshausen, and Trevor Darrell.
\newblock Tent: Fully test-time adaptation by entropy minimization.
\newblock \emph{arXiv preprint arXiv:2006.10726}, 2020.

\bibitem[Wu et~al.(2022)Wu, Wu, Lu, Ju, and Wang]{wu2022style}
Xinyi Wu, Zhenyao Wu, Yuhang Lu, Lili Ju, and Song Wang.
\newblock Style mixing and patchwise prototypical matching for one-shot unsupervised domain adaptive semantic segmentation.
\newblock In \emph{Proceedings of the AAAI Conference on Artificial Intelligence}, volume~36, pages 2740--2749, 2022.

\bibitem[Xiao et~al.(2010)Xiao, Hays, Ehinger, Oliva, and Torralba]{xiao2010sun}
Jianxiong Xiao, James Hays, Krista~A Ehinger, Aude Oliva, and Antonio Torralba.
\newblock Sun database: Large-scale scene recognition from abbey to zoo.
\newblock In \emph{2010 IEEE Computer Society Conference on computer vision and Pattern Recognition}, pages 3485--3492. IEEE, 2010.

\bibitem[Yang et~al.(2022)Yang, Duan, Tran, Xu, Chanda, Chen, Zeng, Chilimbi, and Huang]{yang2022vision}
Jinyu Yang, Jiali Duan, Son Tran, Yi~Xu, Sampath Chanda, Liqun Chen, Belinda Zeng, Trishul Chilimbi, and Junzhou Huang.
\newblock Vision-language pre-training with triple contrastive learning.
\newblock In \emph{Proceedings of the IEEE/CVF Conference on Computer Vision and Pattern Recognition}, pages 15671--15680, 2022.

\bibitem[Zagoruyko and Komodakis(2017)]{attentiondistillation}
Sergey Zagoruyko and Nikos Komodakis.
\newblock Paying more attention to attention: Improving the performance of convolutional neural networks via attention transfer.
\newblock In \emph{ICLR}, 2017.

\bibitem[Zhang et~al.(2023{\natexlab{a}})Zhang, Chao, Dhurandhar, Chen, Tajer, Xu, and Yan]{zhang2023spectral}
Jiajin Zhang, Hanqing Chao, Amit Dhurandhar, Pin-Yu Chen, Ali Tajer, Yangyang Xu, and Pingkun Yan.
\newblock Spectral adversarial mixup for few-shot unsupervised domain adaptation.
\newblock In \emph{Medical Image Computing and Computer Assisted Intervention -- MICCAI 2023}, pages 728--738, Cham, 2023{\natexlab{a}}. Springer Nature Switzerland.

\bibitem[Zhang and Ma(2020)]{detectiondistillation}
Linfeng Zhang and Kaisheng Ma.
\newblock Improve object detection with feature-based knowledge distillation: Towards accurate and efficient detectors.
\newblock In \emph{International Conference on Learning Representations}, 2020.

\bibitem[Zhang et~al.(2020)Zhang, Shi, Shi, Ma, and Bao]{zhang2020task}
Linfeng Zhang, Yukang Shi, Zuoqiang Shi, Kaisheng Ma, and Chenglong Bao.
\newblock Task-oriented feature distillation.
\newblock \emph{Advances in Neural Information Processing Systems}, 33:\penalty0 14759--14771, 2020.

\bibitem[Zhang et~al.(2022)Zhang, Zhang, Fang, Gao, Li, Dai, Qiao, and Li]{zhang2022tip}
Renrui Zhang, Wei Zhang, Rongyao Fang, Peng Gao, Kunchang Li, Jifeng Dai, Yu~Qiao, and Hongsheng Li.
\newblock Tip-adapter: Training-free adaption of clip for few-shot classification.
\newblock In \emph{European Conference on Computer Vision}, pages 493--510. Springer, 2022.

\bibitem[Zhang et~al.(2023{\natexlab{b}})Zhang, Hu, Li, Huang, Deng, Qiao, Gao, and Li]{zhang2023prompt}
Renrui Zhang, Xiangfei Hu, Bohao Li, Siyuan Huang, Hanqiu Deng, Yu~Qiao, Peng Gao, and Hongsheng Li.
\newblock Prompt, generate, then cache: Cascade of foundation models makes strong few-shot learners.
\newblock In \emph{Proceedings of the IEEE/CVF Conference on Computer Vision and Pattern Recognition}, pages 15211--15222, 2023{\natexlab{b}}.

\bibitem[Zhang et~al.(2018)Zhang, Xiang, Hospedales, and Lu]{deepmutuallearning}
Ying Zhang, Tao Xiang, Timothy~M Hospedales, and Huchuan Lu.
\newblock Deep mutual learning.
\newblock In \emph{Proceedings of the IEEE conference on computer vision and pattern recognition}, pages 4320--4328, 2018.

\bibitem[Zhou et~al.(2022{\natexlab{a}})Zhou, Yang, Loy, and Liu]{zhou2021coop}
Kaiyang Zhou, Jingkang Yang, Chen~Change Loy, and Ziwei Liu.
\newblock Learning to prompt for vision-language models.
\newblock \emph{International Journal of Computer Vision}, 130\penalty0 (9):\penalty0 2337--2348, 2022{\natexlab{a}}.

\bibitem[Zhou et~al.(2022{\natexlab{b}})Zhou, Yang, Loy, and Liu]{zhou2022conditional}
Kaiyang Zhou, Jingkang Yang, Chen~Change Loy, and Ziwei Liu.
\newblock Conditional prompt learning for vision-language models.
\newblock In \emph{Proceedings of the IEEE/CVF Conference on Computer Vision and Pattern Recognition}, pages 16816--16825, 2022{\natexlab{b}}.

\bibitem[Zhu et~al.(2023)Zhu, Zhang, He, Zhou, Wang, Zhao, and Gao]{zhu2023not}
Xiangyang Zhu, Renrui Zhang, Bowei He, Aojun Zhou, Dong Wang, Bin Zhao, and Peng Gao.
\newblock Not all features matter: Enhancing few-shot clip with adaptive prior refinement.
\newblock In \emph{Proceedings of the IEEE/CVF International Conference on Computer Vision}, pages 2605--2615, 2023.

\bibitem[Zou et~al.(2019)Zou, Yu, Liu, Kumar, and Wang]{zou2019confidence}
Yang Zou, Zhiding Yu, Xiaofeng Liu, BVK Kumar, and Jinsong Wang.
\newblock Confidence regularized self-training.
\newblock In \emph{Proceedings of the IEEE/CVF International Conference on Computer Vision}, pages 5982--5991, 2019.

\end{thebibliography}
\end{document}


\maketitle

\section{More Dataset Details}
The detailed statistics of each dataset and the corresponding prompt engineering are shown in Table~\ref{dataset details}. We follow~\cite{zhang2022tip,zhou2021coop} and remove the ``BACKGROUND\_Google'' and ``Faces\_easy'' classes from Caltech101. For UCF101, we only take the middle frame of each video for the image encoder. 

\begin{table}[t]
  \begin{minipage}{1.0\textwidth}
   \centering \small
  \setlength{\tabcolsep}{3pt}
  \scalebox{0.82}[0.82]{
  \begin{tabular}{l|ccccp{4cm}}
  \toprule
  \rowcolor{Gray} 
  Dataset &Abbreviation & Class Number & Train Set & Test Set & Prompt Engineering \\
    \midrule
    ImageNet &ImgNet &1,000 &1.28M &50,000 & ``itap of a [class].", 
                        ``a bad photo of the [class]."
                        ``an origami [class].",
                        ``a photo of the large [class].",
                        ``a [class] in a video game.",
                        ``art of the [class].",
                        ``a photo of the small [class]."  \\
                        \midrule
Caltech101&Caltech &100 &4,128&2,465&``a photo of a [class].''  \\
\midrule
DTD&DTD &47&2,820&1,692&``[class] texture.'' \\
\midrule
EuroSAT&ESAT &10&13,500&8,100&``a centred satellite photo of [class].'' \\
\midrule
FGVCAircraf&FGVCA &100&3,334&3,333&``a photo of a [class], a type of aircraft.'' \\
\midrule
Food101&Food &101&50,500&30,300&``a photo of [class], a type of food.''  \\
\midrule
Flowers102&Flower &102&4,093&2,463&``a photo of a [class], a type of flower.'' \\
\midrule
OxfordPets&OxPets &37&2,944&3,669&``a photo of a [class], a type of pet.''  \\
\midrule
SUN397&SUN &397&15,880&19,850&``a photo of a [class].''  \\
\midrule
StandfordCars&StCars &196&6,509&8,041&``a photo of a [class].''  \\
\midrule
UCF101&UCF &101&7,639&3,783&``a photo of a person doing [class].'' \\
\bottomrule
  \bottomrule
  \end{tabular}}
  \end{minipage}
  \begin{minipage}{1.0\textwidth}
  \centering 
  \vspace{0.3em}
  \caption{\footnotesize The detailed dataset statistics and the corresponding handcraft prompts.}
 \label{dataset details}
 \end{minipage}
\end{table}

\section{More Implementation Details}
To generate pseudo-labels for the unlabeled target data, we adhere to the data pre-processing pipeline established by CLIP.  This pipeline involves random cropping, resizing, and horizontal flipping of images.
However, when constructing and fine-tuning the weighted cache model, we employ a more extensive set of data-specific augmentations, as outlined in Table~\ref{data augmentations}. The use of a comprehensive set of data-specific augmentations, in addition to the standard CLIP pre-processing pipeline, is a critical factor in enhancing the effectiveness of the weighted cache model.

\noindent In instances where pseudo-labels cannot be generated for image data points, we utilize the weights of the CLIP classifier itself as image features. This strategic approach ensures that all available data is effectively incorporated into the model's training process, maximizing the utilization of the dataset. 
Furthermore, we draw inspiration from~\cite{zhang2022tip} and implement prompt ensembling for ImageNet using CLIP. This involves combining the outputs from multiple prompts to generate a more robust and accurate representation of the data. In contrast, for the remaining datasets, we employ a single handcrafted prompt specifically designed to capture the unique characteristics of each dataset. 

\noindent In the context of few-shot unlabeled selection, we adopt the methodology outlined in UPL~\cite{huang2022unsupervised}. This method involves generating pseudo-labels for the entire dataset and selecting the top-k confidence samples per class to enrich the training set.
To ensure a fair comparison, we utilize a large CLIP model (ViT-L-14) for generating pseudo-labels in MaPLe~\cite{khattak2023maple} and PromptSRC~\cite{khattak2023self}, maintaining consistency in the experimental setup across different methodologies.
Finally, The hyperparameters $\alpha$ and $\beta$ are set following the values specified in~\cite{zhang2022tip} for consistency and comparability.

\begin{table}[!t]
  \begin{minipage}{1.0\textwidth}
   \centering \small
  \setlength{\tabcolsep}{3pt}
  \scalebox{0.82}[0.82]{
  \begin{tabular}{l|ccccp{4cm}}
  \toprule
  \rowcolor{Gray}
Dataset &Abbreviation & Data Augmentation \\ \midrule
ImageNet &ImgNet & Random Horizontal Flipping   \\
Caltech101&Caltech & Random Horizontal Flipping + Random Affine     \\
DTD&DTD & Random Horizontal Flipping  \\
EuroSAT&ESAT & ColorJitter    \\
FGVCAircraf&FGVCA & Random Horizontal Flipping + Random Affine     \\
Food101&Food & Random Affine     \\
Flowers102&Flower & Random Horizontal Flipping    \\
OxfordPets&OxPets & Random Horizontal Flipping     \\
SUN397&SUN & Random Horizontal Flipping + Random Affine    \\
StandfordCars&StCars & Random Affine    \\
UCF101&UCF & Random Horizontal Flipping + Random Affine  \\ 
\bottomrule
  \bottomrule
  \end{tabular}}
  \end{minipage}
  \begin{minipage}{1.0\textwidth}
  \centering 
  \vspace{0.3em}
  \caption{\footnotesize The detailed data augmentations utilized for each dataset.}
 \label{data augmentations}
 \end{minipage}
\end{table}

\section{More Experimental Results}
\noindent\textbf{Different number of shots: } To assess the effectiveness of NtUA, we manipulated the quantity of unlabeled data. Remarkably, even when utilizing a minimal amount of unlabeled data, specifically 2, 4, or 8 samples per class (refer to Tables~\ref{sota_vit32_2},~\ref{sota_vit32_4}, and~\ref{sota_vit32_8}), NtUA consistently outperformed alternative methodologies.

\begin{table}[t]
  \begin{minipage}{1.0\textwidth}
   \centering \small
  \setlength{\tabcolsep}{3pt}
  \scalebox{0.82}[0.82]{
  \begin{tabular}{c|c|c|c|c|c|c|c|c|c|c|c|c}
  \toprule
  \rowcolor{Gray} 
  \textbf{Methods}  & \textbf{ImgNet} & \textbf{Caltech} & \textbf{DTD} & \textbf{ESAT} & \textbf{FGVCA} & \textbf{Food} & \textbf{Flower} & \textbf{OxPets} & \textbf{SUN} & \textbf{StCars} & \textbf{UCF} & \textbf{Average} \\
    \midrule
    CLIP-ViT-B/32   &63.77 & 91.48 & 44.09 & 45.27 & 19.17 & 80.40 & 66.59& 87.44	& 62.08 & 60.12& 63.47  &  62.17   \\
    \midrule
    UPL &59.06	&91.48	&46.22&	\textbf{54.67}&	6.93	&78.70	&64.68	&82.69&	\textbf{64.33}&	52.92&	63.52	&60.47\\
    UPL$^{*}$  &  59.46&	92.17&	\textbf{46.87}	&48.67	&5.22	&78.55&	66.71	&84.87	&63.76	&54.57	&66.93	&60.71	\\
    LaFTer& 56.90&	85.11	&44.33	&33.89	&16.95	&75.84&	66.42	&80.08	&58.26	&47.26	&60.67&	56.88   \\
    MaPLe &61.26&	90.22	&41.43&	18.21	&17.94	&79.22	&63.05	&79.50	&64.00	&53.77	&63.73	&57.48\\					
    PromptSRC &58.03&	86.90&	43.68&	34.27&	17.37	&76.43	&67.32&	82.37	&60.07	&49.87	&61.38	&57.97	\\
     \rowcolor{LightCyan}
    NtUA (ours)   &\textbf{64.49}	& \textbf{92.90}	&46.81	&52.21	&\textbf{20.70}	&\textbf{80.66}	&\textbf{72.76}&	\textbf{89.62}&	63.13	&\textbf{61.56}	&\textbf{67.86}&	\textbf{64.79}\\
    \midrule
    Supervised &  64.95	&93.75	&56.38	&72.26	&25.8	&80.93	&84.94	&89.26&	66.48&	65.03	&71.27	&70.10\\
  \bottomrule
  \end{tabular}}
  \end{minipage}
  \begin{minipage}{1.0\textwidth}
  \centering 
  \vspace{0.3em}
  \captionsetup{font={footnotesize}}
  \caption{Comparison of NtUA with Five SOTA adaptation methods over 10 widely adopted image classification benchmarks. We leverage CLIP-ViT-B/32 as the backbone model and evaluate performance in a 2-shot setting.}
 \label{sota_vit32_2}
 \end{minipage}
\vspace{1em}
\end{table}

\begin{table}[!t]
  \begin{minipage}{1.0\textwidth}
   \centering \small
  \setlength{\tabcolsep}{3pt}
  \scalebox{0.82}[0.82]{
  \begin{tabular}{c|c|c|c|c|c|c|c|c|c|c|c|c}
  \toprule
  \rowcolor{Gray} 
  \textbf{Methods} & \textbf{ImgNet}& \textbf{Caltech} & \textbf{DTD} & \textbf{ESAT} & \textbf{FGVCA} & \textbf{Food} & \textbf{Flower} & \textbf{OxPets} & \textbf{SUN} & \textbf{StCars} & \textbf{UCF} & \textbf{Average} \\
    \midrule
    CLIP-ViT-B/32   &63.77 & 91.48 & 44.09 & 45.27 & 19.17 & 80.40 & 66.59& 87.44	& 62.08 & 60.12& 63.47  &  62.17   \\
    \midrule
    UPL  &59.68	&92.41&	48.29	&55.89	&16.77&	79.34	&67.24	&82.61	&64.08&	54.06&	64.39	&62.25    \\
    UPL$^{*}$  &60.93&	92.41	&48.17&	50.26	&15.24&	78.58	&72.55&	84.66	&64.15	&57.37	&\textbf{69.13}	&63.04     \\
    LaFTer& 58.67&	88.97&	47.22&	51.73	&17.22	&76.35&	67.97	&84.16&	60.97&	51.24	&63.71&	60.75   \\
    MaPLe  &62.55	&91.32&	37.29&	39.01& 3.15&	79.85	&64.15&	83.84&	63.82&	51.24	&64.47&	58.24   \\					  
    PromptSRC &  60.08	&91.12	&45.15	&39.19&	17.37&	77.16	&68.98	&83.62	&61.92&	53.54&	64.76	&60.26  \\
     \rowcolor{LightCyan}
    NtUA (ours)&   \textbf{65.11}&	\textbf{94.12}	&\textbf{49.76}&	\textbf{61.40}	&\textbf{18.51}	&\textbf{80.85}&	\textbf{73.20}	&\textbf{89.21}	&\textbf{64.30}	&\textbf{62.13}	&67.38	&\textbf{66.00} \\
    \midrule
    Supervised  & 65.83&	94.73	&60.87&	77.2&	28.05&	81.23	&90.26&	89.29&	68.78&	68.25	&75.6&	62.64    \\
  \bottomrule
  \end{tabular}}
  \end{minipage}
  \begin{minipage}{1.0\textwidth}
  \centering 
  \vspace{0.3em}
  \captionsetup{font={footnotesize}}
  \caption{Comparison of NtUA with Five SOTA adaptation methods over 10 widely adopted image classification benchmarks. We leverage CLIP-ViT-B/32 as the backbone model and evaluate performance in a 4-shot setting.}
 \label{sota_vit32_4}
 \end{minipage}
\end{table}

\begin{table}[!t]
  \begin{minipage}{1.0\textwidth}
   \centering \small
  \setlength{\tabcolsep}{3pt}
  \scalebox{0.82}[0.82]{
  \begin{tabular}{c|c|c|c|c|c|c|c|c|c|c|c|c|c}
  \toprule
  \rowcolor{Gray} 
  \textbf{Methods} & \textbf{ImgNet}& \textbf{Caltech} & \textbf{DTD} & \textbf{ESAT} & \textbf{FGVCA} & \textbf{Food} & \textbf{Flower} & \textbf{OxPets} & \textbf{SUN} & \textbf{StCars} & \textbf{UCF} & \textbf{Average} \\
    \midrule
    CLIP-ViT-B/32   &63.77 & 91.48 & 44.09 & 45.27 & 19.17 & 80.40 & 66.59& 87.44	& 62.08 & 60.12& 63.47  &  62.17   \\
    \midrule
    UPL & 61.07&	92.37	&48.46	&58.67	&17.55	&79.61&	67.93&	84.55	&64.94&	54.98	&64.18	&63.12  \\
    UPL$^{*}$   & 61.86	&92.09	&\textbf{52.48}&	52.44	&\textbf{21.42}	&79.19	&75.11&	86.24&	65.53	&60.85&	69.23&	65.13 \\
    LaFTer &59.69	&90.99	&45.98	&50.65	&18.30&	77.76&	69.10&	82.45	&61.72	&53.74	&64.74	&61.37    \\
    MaPLe &62.47&	91.81&	43.74	&28.58	&18.96	&80.37	&65.25&	85.12	&63.72&	55.15	&62.44	&59.78     \\					
    PromptSRC &61.34	&91.24	&46.22&	48.25	&20.61&	78.79	&71.05	&83.59&	62.91	&55.09&	65.05	&62.19 \\
     \rowcolor{LightCyan}
    NtUA (ours) &  \textbf{65.82}	&\textbf{93.71}	&51.95&	\textbf{59.99}	&20.25	&\textbf{81.39}	&\textbf{76.53}&	\textbf{89.59}	&\textbf{65.85}&	\textbf{65.56}&	\textbf{69.60}&	\textbf{67.29 }\\
    \midrule
    Supervised  & 67.23&	94.85&	65.07&	80.35&	33.09	&81.66&	93.18	&89.67&	71.56&	73.10&	79.20	&75.36\\
  \bottomrule
  \end{tabular}}
  \end{minipage}
  \begin{minipage}{1.0\textwidth}
  \centering 
  \vspace{0.3em}
  \captionsetup{font={footnotesize}}
  \caption{Comparison of NtUA with Five SOTA adaptation methods over 10 widely adopted image classification benchmarks. We leverage CLIP-ViT-B/32 as the backbone model and evaluate performance in an 8-shot setting.}
 \label{sota_vit32_8}
 \end{minipage}
 \vspace{1em}
\end{table}

\bibliography{egbib}